%% file: main.tex
\documentclass[10pt,twocolumn,letterpaper]{article}

\usepackage[pagenumbers]{cvpr} % To force page numbers, e.g. for an arXiv version

\usepackage{booktabs} % 优化表格线（CVPR 允许使用）
\usepackage{xcolor} % 颜色支持（箭头颜色）
\usepackage{amssymb} % 提供 LaTeX 原生箭头命令
\usepackage{geometry} % 微调页边距以匹配 CVPR 规范
\usepackage{array} % 增强表格兼容性（核心对齐依赖）
\usepackage{listings}
\newcommand{\upval}[1]{\textsuperscript{\raisebox{0.1ex}{\textcolor{red}{\ensuremath{\uparrow\scriptsize#1}}}}}

\input{preamble}

\definecolor{cvprblue}{rgb}{0.21,0.49,0.74}
\usepackage[pagebackref,breaklinks,colorlinks,allcolors=cvprblue]{hyperref}

\def\paperID{10809} % *** Enter the Paper ID here
\def\confName{CVPR}
\def\confYear{2026}

\title{MAGMA‑Edu: Multi‑Agent Generative Multimodal Framework for Text–Diagram Educational Question Generation}

\author{
Zhenyu Wu \quad
Jian Li\thanks{Corresponding author} \quad
Hua Huang \\
School of Artificial Intelligence, Beijing Normal University \\
{\tt\small 202421081064@mail.bnu.edu.cn}\quad
{\tt\small jli@bnu.edu.cn}\quad
{\tt\small huahuang@bnu.edu.cn}
}

\begin{document}
\maketitle
\input{sec/0_abstract}
\input{sec/1_intro}

\input{sec/2_relatedwork}

\input{sec/3_question_definition}

\input{sec/4_method}
\input{sec/5_c}
\input{sec/6_experiment}
\input{sec/7_conclusion}

{
    \small
    \bibliographystyle{ieeenat_fullname}
    \bibliography{main}
}

\input{sec/X_suppl}

\end{document}

%% file: preamble.tex
\usepackage{tabularx}
\newtheorem{theorem}{Theorem}
\newtheorem{remark}[theorem]{Remark}
\usepackage{xcolor}
\usepackage{listings}
\usepackage{siunitx}
\lstdefinestyle{pythonstyle}{
  language=Python,
  basicstyle=\ttfamily\small,
  keywordstyle=\color{blue}\bfseries,
  stringstyle=\color{teal},
  commentstyle=\color{gray}\itshape,
  numbers=none,
  showstringspaces=false,
  frame=single,
  rulecolor=\color{black},
  breaklines=true,
  tabsize=4
}

\lstdefinelanguage{json}{
    basicstyle=\ttfamily\small,
    numbers=left,
    numberstyle=\tiny,
    stepnumber=1,
    numbersep=5pt,
    showstringspaces=false,
    breaklines=true,
    frame=single,
    backgroundcolor=\color{gray!5},
    literate=
     *{0}{{{\color{numb}0}}}{1}
      {1}{{{\color{numb}1}}}{1}
      {2}{{{\color{numb}2}}}{1}
      {3}{{{\color{numb}3}}}{1}
      {4}{{{\color{numb}4}}}{1}
      {5}{{{\color{numb}5}}}{1}
      {6}{{{\color{numb}6}}}{1}
      {7}{{{\color{numb}7}}}{1}
      {8}{{{\color{numb}8}}}{1}
      {9}{{{\color{numb}9}}}{1}
      {:}{{{\color{punct}{:}}}}{1}
      {,}{{{\color{punct}{,}}}}{1}
      {\{}{{{\color{delim}{\{}}}}{1}
      {\}}{{{\color{delim}{\}}}}}{1}
      {[}{{{\color{delim}{[}}}}{1}
      {]}{{{\color{delim}{]}}}}{1},
}

\definecolor{numb}{rgb}{0.65,0.15,0.15}
\definecolor{punct}{rgb}{0.25,0.25,0.25}
\definecolor{delim}{rgb}{0.0,0.5,0.5}

\usepackage{booktabs}
\usepackage{siunitx}
\usepackage{xcolor}
\usepackage{makecell}
\usepackage{multirow}
\usepackage[table]{xcolor}
\usepackage{tabularx}
\usepackage{adjustbox}

%% file: sec/0_abstract.tex
\begin{abstract}
Educational illustrations play a central role in communicating abstract concepts, yet current multimodal large language models (MLLMs) remain limited in producing pedagogically coherent and semantically consistent educational visuals. 
We introduce MAGMA‑Edu, a self‑reflective multi‑agent framework that unifies textual reasoning and diagrammatic synthesis for structured educational problem generation. 
Unlike existing methods that treat text and image generation independently, MAGMA‑Edu employs a two‑stage co‑evolutionary pipeline: (1) a generation–verification–reflection loop that iteratively refines question statements and solutions for mathematical accuracy, and (2) a code‑based intermediate representation that enforces geometric fidelity and semantic alignment during image rendering. 
Both stages are guided by internal self‑reflection modules that evaluate and revise outputs until domain‑specific pedagogical constraints are met. 
Extensive experiments on multimodal educational benchmarks demonstrate the superiority of MAGMA‑Edu over state‑of‑the‑art MLLMs. 
Compared to GPT‑4o, MAGMA‑Edu improves the average textual metric from 57.01 to 92.31 (+35.3 pp) and boosts image-text consistency (ITC) from 13.20 to 85.24 (+72 pp). 
Across all model backbones, MAGMA‑Edu achieves the highest scores (Avg‑Text 96.20, ITC 99.12), establishing a new state of the art for multimodal educational content generation and demonstrating the effectiveness of self‑reflective multi‑agent collaboration in pedagogically aligned vision–language reasoning.
\end{abstract}

%% file: sec/1_intro.tex
\begin{figure}
    \centering    
    \includegraphics[width=\linewidth]{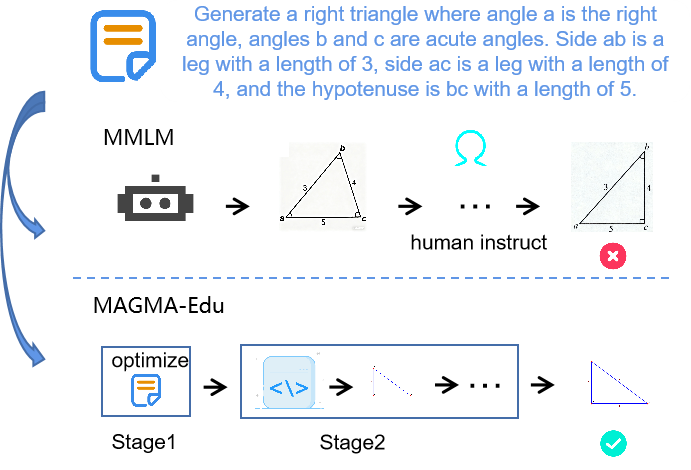}
    \caption{The process of generating geometric images by multimodal large models and MAGMA-Edu. After multiple rounds of human feedback, multimodal large models generate incorrect images, while MAGMA-Edu generates correct images through a two-stage iteration.}
    \label{fig:example}
\end{figure}

\section{Introduction}

High-quality educational resource generation has recently attracted wide attention with the advancement of large language models (LLMs).  
Applications such as personalized learning~\cite{wang2025llmpoweredmultiagentframeworkgoaloriented,10.5555/3737916.3740637,10.1145/3613905.3651122}, automatic Q\&A~\cite{yan-etal-2025-mathagent,Li_Xu_Chang_Wen_2025,tang2025refcritictraininglongchainofthought}, and assignment grading~\cite{Li_Xu_Chang_Wen_2025,yan-etal-2025-mathagent} demonstrate the potential of LLMs to enhance personalization, efficiency, and equity in education.  
However, most existing approaches focus on textual generation, while educational materials are inherently multimodal—images and diagrams are indispensable for conveying abstract concepts and supporting reasoning.  
Thus, \textit{automatically generating accurate and pedagogically sound visual materials remains an unsolved challenge.}

Current multimodal large language models (MLLMs) struggle to meet the requirements of educational image generation due to two fundamental limitations.  
(1) Insufficient text–image semantic alignment: as illustrated in Figure~\ref{fig:example}, even after detailed prompts and multiple manual refinements, generated figures often mislabel geometric elements or distort spatial relations, breaking semantic consistency with the problem text.  
(2) Limited mathematical reliability: due to inherent hallucination in LLMs, single-pass generation often leads to imprecise or logically incorrect results that fail to meet educational standards.  
These deficiencies motivate us to rethink resource generation as a structured, interpretable, and verifiable process rather than an end-to-end black-box output.

To address the above challenges, we propose an alternative solution for multimodal educational resource generation — \texttt{MAGMA-Edu}, a structured multi-agent framework specifically designed for generating mathematical problems with coherent text–diagram pairs.  
Unlike conventional MLLMs that directly synthesize images from text, \texttt{MAGMA-Edu} employs \emph{executable code} as an intermediate representation, ensuring mathematical precision and interpretability.  
The framework follows a two-stage collaborative pipeline:  
\emph{Stage~1} focuses on text refinement and detailed image description generation that align with pedagogical norms;  
\emph{Stage~2} converts these descriptions into verified diagrams through generated drawing code, ensuring consistency between geometric structures and textual semantics.  
Each stage operates under an internal \textit{Generate–Validate–Reflect} cycle, enabling iterative reasoning, self-correction, and cross-modal verification until convergence of textual and visual quality.  
Figure~\ref{fig:example} illustrates how the \texttt{MAGMA‑Edu} process consistently produces correct diagrams even when MLLMs fail after multiple manual adjustments.

On the multimodal educational benchmarks, \texttt{MAGMA‑Edu} achieves the best performance across both textual and visual tasks.  
Compared with GPT‑4o, the average textual score in Stage~1 increases from 57.01 to 92.31, and the image–text consistency (ITC) score in Stage~2 rises from 13.20 to 85.24—an improvement of 72 percentage points—demonstrating the effectiveness of our structured, reflective generation paradigm for educational resources.

Our main contributions are threefold:
\begin{itemize}
    \item We propose \texttt{MAGMA‑Edu}, a training‑free multi‑agent framework enabling multimodal mathematical problem generation with interpretable intermediate code.
    \item We decompose the text–image alignment problem into a two‑stage collaborative optimization process, where each stage ensures high-quality output through an iterative \textit{Generate–Validate–Reflect} mechanism.
    \item We construct a multimodal educational benchmark dataset and verify that \texttt{MAGMA‑Edu} outperforms state-of-the-art multimodal large language models in both textual quality and visual consistency.
\end{itemize}
% \begin{figure*}
%   \centering
%   \begin{subfigure}{0.68\linewidth}
%     \fbox{\rule{0pt}{2in} \rule{.9\linewidth}{0pt}}
%     \caption{An example of a subfigure.}
%     \label{fig:short-a}
%   \end{subfigure}
%   \hfill
%   \begin{subfigure}{0.28\linewidth}
%     \fbox{\rule{0pt}{2in} \rule{.9\linewidth}{0pt}}
%     \caption{Another example of a subfigure.}
%     \label{fig:short-b}
%   \end{subfigure}
%   \caption{Example of a short caption, which should be centered.}
%   \label{fig:short}
% \end{figure*}

%% file: sec/2_relatedwork.tex
\section{RelatedWork}

\textbf{Question Generation} Recent years have seen LLMs’ deep integration with education, offering a new route to advancing educational equity. Relevant research (e.g., personalized teaching \cite{wang2025llmpoweredmultiagentframeworkgoaloriented,10.5555/3737916.3740637,10.1145/3613905.3651122}, intelligent QA \cite{yan-etal-2025-mathagent,Li_Xu_Chang_Wen_2025,tang2025refcritictraininglongchainofthought}) has gained widespread academic attention. Notably, studies are shifting from unimodal to multimodal paradigms: scholars have used multimodal LLMs (MLLM) for automatic exercise grading (solving inefficiencies of traditional manual grading \cite{yan-etal-2025-mathagent}), while others built multimodal Agent-based interaction systems for medical education \cite{wei2024medcomedicaleducationcopilots}. This trend reflects education’s inherent reliance on integrated multimodal information (language, vision, audition).
As a key cross-field research direction, question generation has advanced significantly. Professional educational LLMs (e.g., EduChat\cite{educhat2023}, MudoLLM\cite{MuduoLLM2025}) outperform general-purpose LLMs via strong architectures and context retrieval. Additionally, Prompt Engineering and SFT have optimized model performance \cite{bao2025exploringiterativeenhancementimproving,scarlatos2024improvingautomateddistractorgeneration}. However, existing solutions remain confined to unimodal scenarios, lacking effective mechanisms for image-based logical reasoning exercises. Our core innovation expands the traditional text-only question generation framework into a multimodal paradigm supporting image generation.

\textbf{Image generation} Image is a key core generative capability of multimodal large language models (MLLMs). Mainstream models (e.g., HunyuanImage 3.0 \cite{cao2025hunyuanimage}, Qwen-Image \cite{wu2025qwenimagetechnicalreport}) have achieved human-like style image generation, but text-driven paradigms face key challenges in geometric image generation. Existing MLLMs lack competence in geometric reasoning tasks, struggling to accurately reproduce geometric structures, spatial relationships, and mathematical constraints \cite{deng2025theoremvalidatedreversechainofthoughtproblem,wang2025mathcodervlbridgingvisioncode}. The code-driven generation paradigm is proven effective for geometric images: one approach trains specialized models via image-code alignment datasets to output Python drawing code \cite{wang2025mathcodervlbridgingvisioncode}; another uses a phased framework (mathematical description formalization, key point coordinate calculation, TikZ-based rendering) \cite{wang2025magicgeotrainingfreetextguidedgeometric}. This work takes problem texts and image descriptions from language models as input, automatically generates corresponding Python code, and achieves accurate geometric image rendering.
\textbf{Multi-Agent} Agent is a paradigm that boosts model performance, is widely proven effective \cite{durante2024agentaisurveyinghorizons,wu-etal-2025-agentic} and core to advancing LLM educational applications. Yan et al. \cite{yan-etal-2025-mathagent} used Agent for consistency verification between math problems and images, improving automatic grading accuracy; Wei et al. \cite{wei2024medcomedicaleducationcopilots} integrated models via Agent to build a "multi-role, multi-disciplinary" framework for medical immersive learning; researchers also leveraged Agent for personalized recommendation modules to deliver tailored knowledge \cite{wang2025llmpoweredmultiagentframeworkgoaloriented}.
A key Agent advantage is Self-Reflection, optimizing generation quality—e.g., Liu et al.\cite{liu2024deepseek} designed three Agent roles (student, teacher, principal) for Socratic QA. Drawing on Agent’s "domain adaptation" and "self-reflection" strengths, this study introduces Agent to question generation, aiming to automatically produce high-quality multimodal questions.

%% file: sec/3_question_definition.tex
\begin{figure*}[htbp]
    \centering    
    \includegraphics[width=\linewidth]{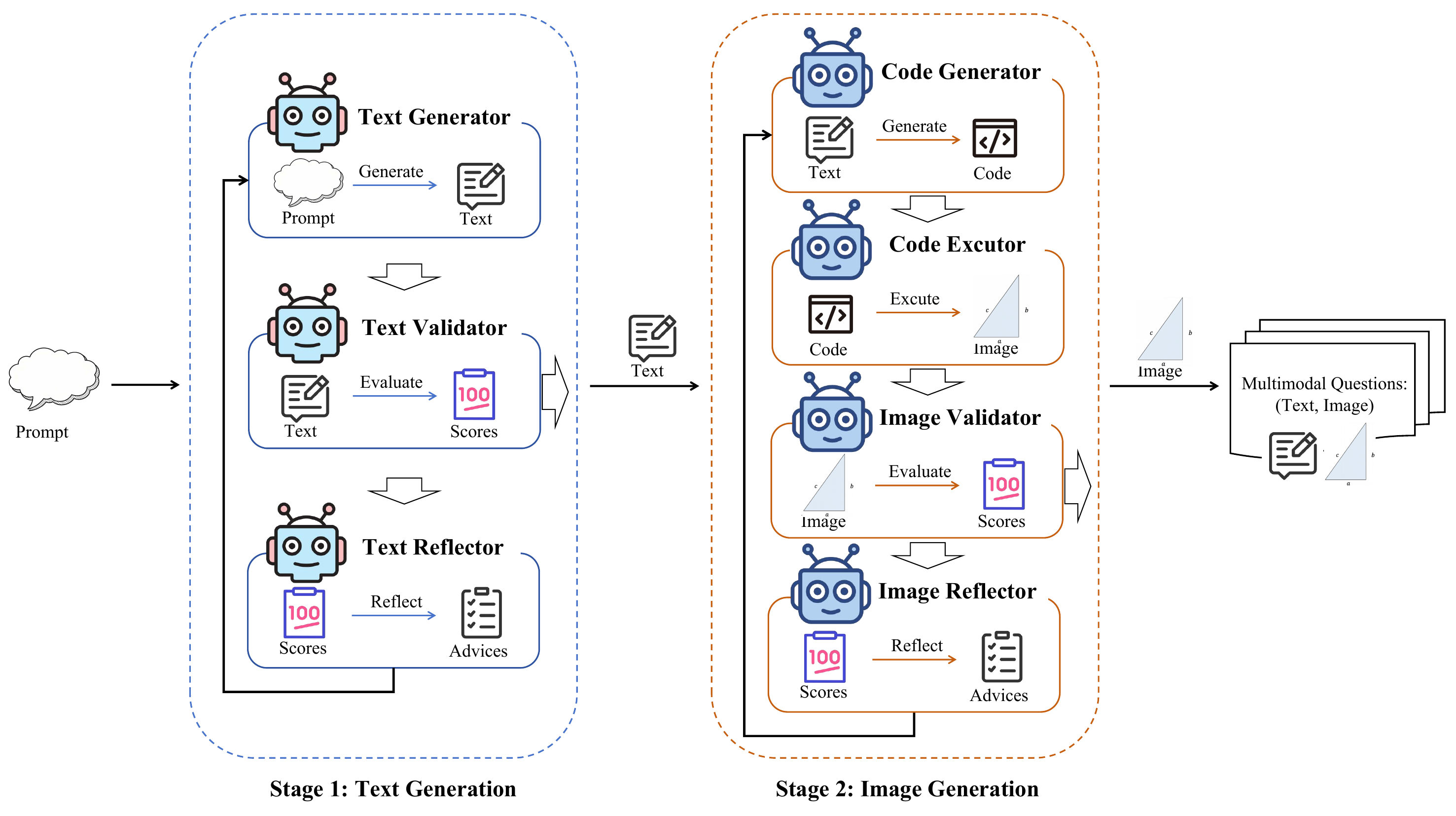}
    \caption{Detailed workflow of the proposed \texttt{MAGMA‑Edu} framework. Stage 1 (\textit{Text Generation}) employs three collaborative agents—Text Generator, Text Validator, and Text Reflector—to iteratively produce, evaluate, and refine problem statements from a given prompt. Stage 2 (\textit{Image Generation}) mirrors this process with Code Generator, Code Executor, Image Validator, and Image Reflector agents, which translate verified text into executable drawing code and refine it into accurate, interpretable diagrams. Both stages form a closed‑loop multimodal optimization system that outputs pedagogically aligned text–image pairs as final questions.}
    \label{fig:framework_stages}
\end{figure*}

\section{Problem Definition}

We define the educational visual question generation task as a systematic multimodal generation problem. 
Given a concise textual description of a knowledge point and task constraints, the goal is to automatically produce a well-structured educational problem --- including 
a question text, a correct and pedagogically coherent solution, and an accompanying geometric diagram that satisfies formal mathematical and visual constraints.

The input to a visual question generation system is a structured instructional representation:
\begin{equation*}
    \mathcal{I} = \{k, s, r\},
\end{equation*}
where $k$ denotes the given \textit{knowledge point}, such as ``Pythagorean theorem''; $s$ denotes the \textit{subject and grade level}, e.g., ``middle school geometry''; and $r$ represents the \textit{diagram and parameters requirements}, specifying geometric entities and conditions to be visualized.

\noindent
For example, $\mathcal{I}$ =\{
  ``Pythagorean theorem'',
  ``junior high geometry'',
  ``a right triangle with legs 3\,cm and 4\,cm; find the hypotenuse''\}.

\noindent
The expected output of visual question generation is a multimodal educational problem instance:
\begin{equation*}
    \mathcal{O} = \{T, G\}.
\end{equation*}
The output $\mathcal{O}$ consists of two parts: the textual component $T$ and the graphic component $G$.
Note that, $T = \{t_q, t_a, t_e\}$, where:
\begin{itemize}
    \item $t_q$: the question stem, e.g.,
    \textit{``In the right triangle $ABC$ shown in the figure, the legs $AB = 3\,\text{cm}$ and $AC = 4\,\text{cm}$.
    Find the length of the hypotenuse $BC$.''}
    \item $t_a$: the final numeric answer, $BC = 5\,\text{cm}$.
    \item $t_e$: the explanation,
    \textit{``By the Pythagorean theorem, $BC^2 = AB^2 + AC^2 = 3^2 + 4^2 = 25$, 
    hence $BC = 5\,\text{cm}$.''}
\end{itemize}
$G$ denotes the corresponding illustrative content, which visualizes the underlying geometric structure and remains semantically aligned with $T$. 
To preserve generality, $G$ can be produced by the following ways:
\begin{itemize}
    \item Direct multimodal synthesis: generated by a multimodal large language model (MLLM) that directly produces the illustrative image.
    \item Program‑based rendering: generated from an intermediate code representation and rendered deterministically using tools such as \texttt{Matplotlib} or TikZ.
\end{itemize}

The overall visual question generation system learns a bijective mapping between the textual educational intent $\mathcal{I}$ and the resulting multimodal question, formally:
\begin{equation}
    (T, G) = \mathcal{F}(\mathcal{I}, \Theta),
\end{equation}
where $\Theta$ denotes the parameters of the multimodal question generation system $\mathcal{F}$ optimized to produce coherent, valid, and visually consistent problems.

The goal of the visual question generation system is to generate an output $\mathcal{O} = (T, G)$ that maximizes overall quality 
in terms of textual accuracy, visual fidelity, and cross‑modal semantic consistency. 
Formally, the optimization objective can be written as:
\begin{equation}
    \label{eq.objective}
    \max_{\Theta} \;
    \mathbb{E}_{\mathcal{I}} \big[ \Phi(T, G) \big],
    \quad \text{where } (T, G) = \mathcal{F}(\mathcal{I}; \theta).
\end{equation}
where the multimodal evaluation $\Phi$ is defined as:
\begin{equation*}
    \begin{aligned}
        \Phi(T, G) =
            &\alpha \cdot Q_{\text{text}}(T)
            + \beta \cdot Q_{\text{image}}(G) \\
            &+ \gamma \cdot \text{Consistency}(T, G),
    \end{aligned}
\end{equation*}
and $\alpha$, $\beta$, and $\gamma$ are non-negative weights that balance the three dimensions of quality.
Specifically:
\begin{itemize}
    \item $Q_{\text{text}}(T)$ evaluates textual quality such as formula correctness, reasoning validity, and clarity;
    \item $Q_{\text{image}}(G)$ measures geometric precision, clear labeling, and visual readability of the diagram;
    \item $\text{Consistency}(T, G)$ evaluates cross‑modal correspondence, ensuring that all textual and quantitative details are correctly rendered in the diagram.
\end{itemize}

\begin{remark}
    The formulation in Eq. \eqref{eq.objective} emphasizes that the visual question generation system is not merely a text‑to‑image task, but a multimodal optimization problem in which textual reasoning and visual synthesis must be jointly optimized under explicit educational and geometric constraints.
\end{remark}

%% file: sec/4_method.tex
\section{Methodology}

The objective of the educational visual question generation task is to formulate an optimal multimodal generation function
$\mathcal{F}^{*}$ that maximizes the overall quality metric according to Eq.~\eqref{eq.objective} by adaptively configuring system parameters 
$\Theta$ (e.g., agent prompts, iteration thresholds, and evaluation weights).
Unlike conventional neural training, $\Theta$ represents meta‑level control variables governing the behaviors and interactions of collaborating agents rather than gradient‑learnable model weights. 
Accordingly, the optimization of $\Theta$ is performed through rule‑based and heuristic adaptation, instead of gradient back‑propagation.

To achieve this objective without any task‑specific fine‑tuning, we propose \texttt{MAGMA‑Edu}—a \textit{Multi‑Agent Generative Multimodal Architecture for Education}. 
It is a general, extensible, and training‑free multi-agent framework for educational multimodal question generation. 
Departing from conventional single‑model or single‑prompt paradigms, 
\texttt{MAGMA‑Edu} orchestrates specialized language and vision agents that collaborate across two iterative stages to jointly produce high‑quality text–image question pairs.

\texttt{MAGMA‑Edu} is organized as a two‑stage collaborative pipeline. 
{Stage 1} performs text generation and refinement, producing pedagogically sound question text 
together with a detailed image description that specifies the visual context. 
{Stage 2} translates this description into executable drawing code and generates a validated diagram 
that is geometrically and semantically consistent with the textual problem. 
Each stage operates through an internal \textit{Generate–Validate–Reflect} loop, 
enabling the agents to iteratively reason, verify, and self‑correct 
until convergence of textual and visual quality or reaching the maximum number of iterations.

For illustration, consider a geometry question based on the 
{Pythagorean theorem}: 
\textit{``In a right triangle with legs of 3\,cm and 4\,cm, find the length of the hypotenuse.''}

\subsection{Stage~1: Text Generation and Reflective Refinement}

Stage~1 focuses on optimizing the textual component quality $Q_{\text{text}}(T)$, 
where $T = \mathcal{F}_{\text{text}}(k, s, r)$.
Its objective is to produce problem statements that are mathematically valid, 
logically coherent, and pedagogically aligned with the target learning objectives. 
This stage emphasizes linguistic precision, conceptual rigor, and instructional clarity.

We introduce a multi‑agent reflective refinement mechanism in which three specialized agents collaborate within a self‑improving feedback loop:

\begin{itemize}
  \item \textbf{Text Generator Agent} --- 
  synthesizes an initial multimodal problem instance from structured instructional input 
  $\mathcal{I} = \{\,k=\text{``Pythagorean theorem''},\, s=\text{``junior‑high geometry''},\, r=\text{``legs 3 cm, 4 cm''}\,\}$, 
  constructing a semantically grounded and contextually relevant problem draft.
  
  \item \textbf{Text Validator Agent} --- 
  conducts automated evaluation across multiple quality dimensions—including mathematical correctness, linguistic fluency, structural completeness, and pedagogical soundness—using the formalized multi‑dimensional metrics defined in Section \ref{sec.evaluation}.
  
  \item \textbf{Text Reflector Agent} --- 
  aggregates the validator’s diagnostic feedback, abstracts high‑level revision cues, 
  and translates them into actionable textual refinements, 
  enabling iterative self‑correction and reasoning traceability.
\end{itemize}

The Generator Agent first produces a structured JSON representation of the educational problem:
\begin{lstlisting}[language=json, basicstyle=\ttfamily\small, breaklines=true, frame=single]
{
  "subject": "Mathematics",
  "grade_level": "Junior High",
  "knowledge_point": "Pythagorean theorem",
  "question_stem": "In the right triangle ABC shown in the figure,
     AB = 3 cm and AC = 4 cm. Find BC.",
  "image_description": "A right triangle with AB=3 cm, AC=4 cm,
     right angle at A, labeled A, B, C.",
  "answer": "5 cm",
  "analysis": "By the Pythagorean theorem,
     BC^2 = AB^2 + AC^2 = 9 + 16 = 25, so BC = 5 cm."
}
\end{lstlisting}

If the Text Validator Agent detects deficiencies—for example, an incomplete theorem formulation—it issues a structured feedback message:
\begin{lstlisting}[language=json, basicstyle=\ttfamily\small, breaklines=true, frame=single]
{
 "status": "revision_needed",
 "feedback": [
   {"error_type": "formula_error",
    "suggestion": "Include full expression BC^2 = AB^2 + AC^2."}
 ]
}
\end{lstlisting}

The Text Reflector Agent analyzes multi‑source feedback 
(e.g., linguistic coherence, factual correctness, and pedagogical alignment) 
and formulates a targeted update signal $\Delta_T^{(i)}$ 
to guide the next refinement round. 
Formally, the textual representation at iteration $(i{+}1)$ is updated as:
\begin{equation}
    T^{(i+1)} = 
        \mathcal{F}_{\text{text}}\big(k, s, r, \Delta_T^{(i)}\big),
\end{equation}
where $\mathcal{F}_{\text{text}}(\cdot)$ denotes the text‑generation 
and refinement function parameterized by the instructional knowledge $k$, 
semantic context $s$, and external review feedback $r$.  
The term $\Delta_T^{(i)}$ encapsulates the adaptive adjustment derived from the 
Reflector Agent, representing magnitude of textual improvement at iteration $i$.

This iterative process constitutes a form of closed‑loop self‑supervised optimization, 
where the system continuously evaluates and refines its own outputs 
until the generated text satisfies predefined quality criteria—%
such as semantic precision, factual validity, and didactic appropriateness—%
yielding a refined and verified statement $T^{*}$.

Distinct from conventional one‑shot text generation pipelines, 
this stage embodies a self‑refining, agent‑centric paradigm that 
tightly integrates instructional semantics, formal verification, 
and iterative reflection. Such a design substantially enhances the 
factual consistency and pedagogical reliability of the generated questions, 
while establishing a scalable foundation for multimodal 
educational content synthesis.

\begin{remark}
To ensure computational efficiency and prevent infinite recursion, 
the iterative loop is executed under a bounded optimization scheme. 
Specifically, the refinement process terminates once either  
the textual quality metric $Q_{\text{text}}(T^{(i)})$  
meets or exceeds a predefined threshold $\tau_{\text{text}}$,  
indicating satisfactory linguistic and pedagogical adequacy,  
or the maximum iteration count $I_{\max}$ is reached:
\begin{equation}
\text{Stop if}\;
Q_{\text{text}}(T^{(i)}) \ge \tau_{\text{text}}
\;\;\text{or}\;\;
i \ge I_{\max}.
\end{equation}
This constraint effectively prevents uncontrolled resource consumption 
while guaranteeing convergence to a textually sound and pedagogically 
appropriate representation $T^{*}$.
\end{remark}

\subsection{Stage~2: Programmatic Diagram Generation and Reflective Correction}

Based on the text $T^*$ and input $r$, Stage~2 aims to optimize the visual component $Q_{\text{visual}}(G)$ 
together with the cross‑modal consistency term $\mathrm{Consistency}(T^{*}, G)$, where $G = 
    \mathcal{F}_{\text{vision}}\big(T^{*}, r\big),$
producing a geometrically accurate and semantically interpretable diagram. 
Given the verified problem text $T^{*}$ from Stage~1, 
this stage focuses on translating linguistic descriptions into  
executable graphical representations that faithfully preserve the original semantics.  

Four specialized agents collaborate within a closed‑loop refinement pipeline 
to ensure code validity, geometric correctness, and semantic consistency:

\begin{itemize}
  \item \textbf{Code Generator Agent} --- 
  converts the finalized text description $T^{*}$ into executable drawing code 
  (e.g., in \texttt{Python/Matplotlib}), encoding spatial relations, labels, and dimensions.

  \item \textbf{Code Executor Agent} --- 
  compiles and runs the generated script to render the image $G$,  
  while monitoring runtime stability and graphical completeness.

  \item \textbf{Image Validator Agent} —  
  parses the drawing code and evaluates both code–text and image–description alignment through multimodal reasoning (e.g., OCR‑based comparison of embedded textual labels), yielding the quality metrics $\{Q_{\text{syntax}}, Q_{\text{visual}}, Q_{\text{align}}\}$ as defined in Section~\ref{sec.evaluation}.

  \item \textbf{Image Reflector Agent} --- 
  aggregates all diagnostic feedback and synthesizes a corrective update $\Delta_G^{(i)}$ to guide the next iteration of visual refinement. 
\end{itemize}

An illustrative output from the Code Generator Agent is shown below:

\begin{lstlisting}[style=pythonstyle]
import matplotlib.pyplot as plt
plt.plot([0,3,0,0],[0,0,4,0])  # draw right triangle
plt.text(0,0,'A'); plt.text(3,0,'B'); plt.text(0,4,'C')
plt.text(1.5,-0.4,'3 cm'); plt.text(-0.6,2,'4 cm')
plt.text(1.6,1.8,'5 cm'); plt.axis('equal')
plt.savefig('triangle.svg')
\end{lstlisting}

If the Validator Agent detects visual or semantic discrepancies  
(e.g., missing labels, misplaced right‑angle markers, or inconsistent scaling),  
it issues a structured feedback package $\Delta_G^{(i)}$.  
The Code Generator Agent then incorporates this feedback to update the code,  
yielding an improved visual representation through an iterative correction process: 
\begin{equation}
    G^{(i+1)} = 
    \mathcal{F}_{\text{vision}}\big(T^{*}, r, \Delta_G^{(i)}\big),
\end{equation}
where $\mathcal{F}_{\text{vision}}(\cdot)$ represents the vision‑generation  
and correction function parameterized by the input text $T^{*}$,  
rendering context $r$, and iterative visual feedback $\Delta_G^{(i)}$. 

The process continues until all validation criteria are met  
(i.e., $Q_{\text{visual}}(G^{(i)}) \ge \tau_{\text{visual}}$ or $i \ge I_{\max}$),  
resulting in a verified, consistent diagram $G^{*}$.  
Through this reflective, program‑driven refinement cycle,  
Stage~2 ensures that each generated diagram attains visual accuracy,  
structural clarity, and precise alignment with its textual specification.  
Together with Stage~1, it forms the end‑to‑end multimodal synthesis  
and verification architecture of \texttt{MAGMA‑Edu}.

\begin{table*}[!t]
\centering
\caption{Performance comparison between single‑agent MLLMs and the proposed MAGMA‑Edu on textual (Avg‑Text, the arithmetic mean of all six textual metrics) and visual (ITC) metrics. The best result in each column is highlighted in green, and relative improvements (\upval{+}) of MAGMA‑Edu over the corresponding MLLM baselines are reported.}
\label{tab:model_performance}
\setlength{\tabcolsep}{4pt}   % 稍微缩小列间距
\renewcommand{\arraystretch}{1.1}

\begin{adjustbox}{max width=\textwidth}  % 自动按页面宽度缩放
\begin{tabular}{llcccccccc}
\toprule
\textbf{Category} & \textbf{Model} 
  & \multicolumn{7}{c}{\textbf{Textual Metrics}} 
  & \textbf{Visual Metric} \\ 
\cmidrule(lr){3-9} \cmidrule(lr){10-10}
 &  & \textbf{UO} & \textbf{LR} & \textbf{QF} & \textbf{AA} & \textbf{CA} & \textbf{IDQ} & \textbf{Avg-Text} & \textbf{ITC} \\
\midrule

\multirow{2}{*}{\textbf{Single Agent}} 
 & GPT-4o & 32.47 & 58.67 & 63.47 & 56.83 & 67.16 & 63.47 & 57.01 & 13.20 \\
 & Nano-Banana & 99.63 & 99.63 & 89.30 & 96.68 & 83.39 & 100.00 & 94.77 & 15.90 \\
 % & GPT-5 & 99.63 & 100.00 & 84.87 & 96.68 & 81.18 & 100.00 & 93.73 & 0.00 \\
 % & DeepSeek-R1 & 99.26 & 99.26 & 87.08 & 97.05 & 84.87 & 100.00 & 94.59 & 0.00 \\
 % & Gemini 2.5 Pro & 97.81 & 100.00 & 88.60 & 97.37 & 93.42 & 100.00 & \cellcolor{green!20}\textbf{96.20} & 0.00 \\
\midrule

\multirow{5}{*}{\textbf{MAGMA-Edu}} 
 & GPT-4o & 98.89 & 97.42 & 89.67 & 84.13 & 83.76 & 100.00 & 92.31\upval{+35.30} & 85.24\upval{+72.04} \\
 & Nano-Banana & 100.00 & 100.00 & 89.30 & 97.05 & 87.08 & 100.00 & 95.57\upval{+0.80} & 97.05\upval{+81.15} \\
 & GPT-5 & 99.63 & 100.00 & 91.51 & 96.68 & 81.92 & 100.00 & 94.96 & 95.20 \\
 & DeepSeek-R1 & 99.26 & 99.26 & 94.46 & 96.31 & 85.98 & 99.63 & 95.82 & 94.10 \\
 & Gemini 2.5 Pro & 97.81 & 99.56 & 86.84 & 98.68 & 94.30 & 100.00 & \cellcolor{green!20}\textbf{96.20} & \cellcolor{green!20}\textbf{99.12} \\
\bottomrule
\end{tabular}
\end{adjustbox}
\end{table*}

\subsection{Discussion and Innovation Highlights}

\texttt{MAGMA‑Edu} redefines multimodal educational content generation as agent-based reflective reasoning, rather than direct text-to-image translation. Its innovations lie in three interlocking dimensions:

\begin{itemize}
  \item \textbf{General Collaborative Multi‑agent Optimization.}  
  A domain-agnostic coordination architecture where textual, visual, and reflective heterogeneous agents collaboratively optimize the joint objective $\Phi(T,G)$ without model-specific retraining. It enables cross-disciplinary generalization and adaptive multimodal cognition in education.

  \item \textbf{Programmatic Intermediate Representation.}  
  Executable code serves as a cross-modal lingua franca, bridging symbolic reasoning and visual perception. It enforces geometric constraints, guarantees interpretability/reproducibility, and transforms diagram synthesis into verifiable algorithmic reasoning.

  \item \textbf{Iterative Cross‑modal Self‑reflection.}  
  Recursive generation-validation-reflection cycles emulate human learning (conceptualizing, sketching, verifying, revising), achieving self-consistent co-evolution of language and vision. This operationalizes self-regulated multimodal learning, paving the way for interpretable autonomous cross-modal intelligence.
\end{itemize}
In summary, \texttt{MAGMA‑Edu} advances multimodal generation 
from a descriptive process to an \textit{explanatory reasoning framework}.  
By coupling natural‑language understanding with programmatic visual synthesis 
through reflective agent collaboration,
it not only ensures factual and pedagogical reliability, 
but also contributes a new paradigm of cross‑modal self‑optimization—%
a scalable route toward transparent and verifiable AI 
for educational knowledge construction.
% \texttt {MAGMA‑Edu} advances multimodal generation from descriptive to explanatory reasoning. Via coupling natural-language understanding with programmatic visual synthesis and reflective agent collaboration, it ensures factual and pedagogical reliability while offering a cross-modal self-optimization paradigm—enabling scalable, transparent, verifiable AI for educational knowledge construction.

%% file: sec/5_c.tex
\section{Evaluation Metrics}
\label{sec.evaluation}
To comprehensively and objectively evaluate the proposed \texttt{MAGMA‑Edu} framework, 
we extend and adapt five textual evaluation metrics originally developed for purely text‑based educational question generation~\cite{zhou2025answers} 
to our \textit{multimodal} setting. 
A concise overview is presented below, while detailed definitions and illustrative examples 
are provided in the Appendix.

Six metrics are employed to evaluate the textual quality of the generated questions:
\begin{enumerate*}[label=(\arabic*), itemindent=0em, leftmargin=2em]
    \item \textbf{User Orientation (UO)} — Evaluates whether the generated question satisfies the instructional requirements specified in the system input $\mathcal{I}$.
    \item \textbf{Language Readability (LR)} — Ensures grammatical fluency and the absence of corrupted characters or non‑standard symbols in the generated content.
    \item \textbf{Question Feasibility (QF)} — Assesses the rationality and pedagogical appropriateness of the question stem and its associated image information.
    \item \textbf{Accurate Analysis (AA)} — Examines the logical soundness and coherence of the reasoning presented in the generated explanation or solution.
    \item \textbf{Correct Answer (CA)} — Verifies the numerical or symbolic correctness of the final answer derived for the question.
    \item \textbf{Image Description Quality (IDQ)} — Evaluates whether the image description accurately captures the intended textual requirements and visual context.
\end{enumerate*}

Beyond textual quality, overall multimodal alignment is quantified through an additional integrated metric:
\begin{enumerate*}[label=(\arabic*), itemindent=0em, leftmargin=2em]
    \item \textbf{Image–Text Consistency (ITC)} — A question is considered valid if it successfully passes all three verification stages: code quality $\{Q_{\text{syntax}}\}$, code‑text alignment $\{Q_{\text{align}}\}$, 
    and multimodal reasoning $\{Q_{\text{visual}}\}$.
\end{enumerate*}

%% file: sec/6_experiment.tex
\section{Experiments}

\begin{table*}[!t]
\centering
\caption{Ablation study of the proposed MAGMA‑Edu framework on textual (Avg‑Text) and visual (ITC) metrics. Relative improvements (\upval{+}) from Stage 1, Stage 2, and MAGMA‑Edu over each baseline are shown.}
\label{tab:ablation}
\setlength{\tabcolsep}{4pt}
\renewcommand{\arraystretch}{1.1}

\begin{adjustbox}{max width=\textwidth}
\begin{tabular}{llcccccccc}
\toprule
\textbf{Category} & \textbf{Model}
  & \multicolumn{7}{c}{\textbf{Textual Metrics}} 
  & \textbf{Visual Metric} \\ 
\cmidrule(lr){3-9} \cmidrule(lr){10-10}
 &  & \textbf{UO} & \textbf{LR} & \textbf{QF} & \textbf{AA} & \textbf{CA} & \textbf{IDQ} & \textbf{Avg‑Text} & \textbf{ITC} \\
\midrule

% ---------- Multimodal LLMs ----------
\multirow{8}{*}{\textbf{MLLMs}} 
 & GPT‑4o & 32.47 & 58.67 & 63.47 & 56.83 & 67.16 & 63.47 & 57.01 & 13.20 \\
 & \hspace{1em}+Stage 1 & 98.89 & 97.42 & 89.67 & 84.13 & 83.76 & 100.00 & 92.31\upval{+35.30} & 14.90\upval{+1.70} \\
 & \hspace{1em}+Stage 2 & 32.47 & 58.67 & 63.47 & 56.83 & 67.16 & 63.47 & 57.01\upval{+0.00} & 75.65\upval{+62.45} \\
 & \hspace{1em}+MAGMA‑Edu & 98.89 & 97.42 & 89.67 & 84.13 & 83.76 & 100.00 & 92.31\upval{+35.30} & 85.24\upval{+72.04} \\ \cmidrule(lr){2-10}
 & Nano‑Banana & 99.63 & 99.63 & 89.30 & 96.68 & 83.39 & 100.00 & 94.77 & 15.90 \\
 & \hspace{1em}+Stage 1 & 100.00 & 100.00 & 89.30 & 97.05 & 87.08 & 100.00 & 95.57\upval{+0.80} & 17.80\upval{+1.90} \\
 & \hspace{1em}+Stage 2 & 99.63 & 99.63 & 89.30 & 96.68 & 83.39 & 100.00 & 94.77\upval{+0.00} & 87.45\upval{+71.55} \\
 & \hspace{1em}+MAGMA‑Edu & 100.00 & 100.00 & 89.30 & 97.05 & 87.08 & 100.00 & 95.57\upval{+0.80} & 97.05\upval{+81.15} \\

\midrule

% ---------- Text-only LLMs ----------
\multirow{6}{*}{\textbf{LLMs}} 
 & GPT‑5 & 99.63 & 100.00 & 84.87 & 96.68 & 81.18 & 100.00 & 93.73 & 0.00 \\
 & \hspace{1em}+MAGMA‑Edu & 99.63 & 100.00 & 91.51 & 96.68 & 81.92 & 100.00 & 94.96\upval{+1.23} & 95.20\upval{+95.20} \\ \cmidrule(lr){2-10}
 & DeepSeek‑R1 & 99.26 & 99.26 & 87.08 & 97.05 & 84.87 & 100.00 & 94.59 & 0.00 \\
 & \hspace{1em}+MAGMA‑Edu & 99.26 & 99.26 & 94.46 & 96.31 & 85.98 & 99.63 & 95.82\upval{+1.23} & 94.10\upval{+94.10} \\ \cmidrule(lr){2-10}
 & Gemini 2.5 Pro & 97.81 & 100.00 & 88.60 & 97.37 & 93.42 & 100.00 & 96.20 & 0.00 \\ 
 & \hspace{1em}+MAGMA‑Edu & 97.81 & 99.56 & 86.84 & 98.68 & 94.30 & 100.00 & 96.20\upval{+0.00} & 99.12\upval{+99.12} \\
\bottomrule
\end{tabular}
\end{adjustbox}
\end{table*}

To comprehensively verify the effectiveness of the proposed \texttt{MAGMA‑Edu} framework, 
we design three experimental studies:  
(1)~a \textit{comparison experiment} with existing multimodal large language models (MLLMs),  
(2)~an \textit{ablation experiment} to investigate the contribution of each system component, and  
(3)~a \textit{knowledge‑point coverage experiment} to examine the breadth and balance 
of generated question topics.

\subsection{Experimental Setting}

% A multimodal mathematics dataset was constructed for K–12 education, 
% covering both junior and senior high school curricula. 
% The dataset includes $78$ fine‑grained multimodal knowledge points such as plane geometry, analytic geometry, solid geometry, trigonometric functions, function graphs, and composite function visualization.  
% Each knowledge point integrates mathematical concepts with corresponding diagram descriptions.

% For each subtopic, natural‑language prompts were manually crafted to simulate 
% realistic teacher instructions in alignment with the input specifications of the \texttt{MAGMA‑Edu} system.  
% For instance:  
% \textit{“I am a seventh‑grade mathematics teacher preparing a unit test.  
% Please create an applied problem that examines the properties of complementary and supplementary angles, includes a figure, and requires students to reason based on the diagram.”}  
% For each knowledge point, the system was prompted to generate 
% five distinct multimodal questions, resulting in a total of 
% $390$ candidate problems that combine textual and visual components.
\textbf{Dataset.} A multimodal K–12 mathematics dataset is built, covering junior and senior high curricula. It includes 78 fine-grained multimodal knowledge points (e.g., plane geometry, analytic geometry, solid geometry, trigonometric functions, function graphs, composite function visualization), each integrating mathematical concepts with corresponding diagram descriptions.
Manually crafted natural-language prompts simulate realistic teacher instructions, aligning with \texttt{MAGMA-Edu}'s input requirements. Example: “I’m a seventh-grade math teacher preparing a unit test. Create an applied problem on complementary/supplementary angle properties with a figure, requiring reasoning based on the diagram.” Each knowledge point prompts the system to generate 5 distinct multimodal questions, resulting in 390 candidate problems combining text and corresponding visual diagrams.
The default maximum iteration count is $I_\text{max} = 5$.

% Among publicly available multimodal large language models (MLLMs), 
% only a few currently support both text and diagram generation. 
% We therefore select two representative baselines for comparison: GPT‑4o \cite{hurst2024gpt} and Gemini2.5‑Flash‑Img (abbreviated as {Nano‑Banana}) \cite{nano-banana2025}. 
 
% In addition, state‑of‑the‑art LLMs without native visual generation capabilities, e.g. GPT‑5 \cite{openai2025gpt5}, Gemini 2.5 Pro \cite{gemini2025}, and DeepSeek‑V3.1 \cite{liu2024deepseek}, are evaluated solely on textual metrics.  
% When integrated into the proposed framework, their performance on image‑related criteria is also assessed to examine the effectiveness of our multimodal extension.

% The default agent backbone of \texttt{MAGMA‑Edu} is DeepSeek‑V3.1.  
% In ablation settings, different models are used to replace either the text‑generation agent or the code‑generation agent, while all other configurations remain unchanged.  
% The verification procedure follows the same methodology as in Stage 1 and Stage 2.  

\noindent\textbf{Baseline Models.} 
Among publicly available multimodal large language models (MLLMs), few currently support both text and diagram generation. 
We select two representative models for comparison: GPT‑4o~\cite{hurst2024gpt} and \texttt{Gemini2.5‑Flash‑Img} (abbreviated as Nano‑Banana)~\cite{nano-banana2025}, which can generate both text and image. 
State‑of‑the‑art LLMs without native visual generation capabilities 
(e.g., GPT‑5~\cite{openai2025gpt5}, \texttt{Gemini~2.5~Pro}~\cite{gemini2025}, and \texttt{DeepSeek‑V3.1}~\cite{liu2024deepseek}) 
are first evaluated on textual metrics. 
When integrated into our framework, their image‑related performance is further assessed to validate the effectiveness of the proposed multimodal extension.  

\texttt{MAGMA‑Edu} uses \texttt{DeepSeek‑V3.1} as the default agent backbone. 
For ablation studies, we replace either the text‑generation or code‑generation agent with alternative models while keeping other settings fixed, and evaluate them following the same methodology as Stage~1 and Stage~2.

\subsection{Comparison Experiments}
Table~\ref{tab:model_performance} presents a comprehensive performance comparison across six textual metrics and one visual metric (ITC) for all evaluated models. 
Overall, \texttt{MAGMA‑Edu} consistently improves both textual and visual outcomes compared with the single‑agent baselines.  

\textbf{Textual Performance.}
Among single‑agent models, \texttt{Gemini~2.5~Pro} achieves the highest average textual score (96.20), surpassing other baselines such as Nano‑Banana (94.10) and DeepSeek‑R1 (94.25). 
This demonstrates \texttt{Gemini~2.5~Pro}'s strong language understanding and generation ability. 
After integrating \texttt{MAGMA‑Edu}, all MLLMs exhibit clear gains in textual metrics. 
For instance, GPT‑4o improves from 57.51 to 92.65 (+35.14), while Nano‑Banana further increases from 94.77 to 95.57. 
These results validate the effectiveness of \texttt{MAGMA‑Edu} in enhancing textual comprehension and expression quality.

\textbf{Visual performance.}
In terms of the visual ITC metric, all single-agent models perform poorly (e.g., GPT-4o = 13.20, Nano-Banana = 15.90). After applying MAGMA-Edu, the visual alignment improves dramatically: GPT-4o increases to 85.24, GPT-5 reaches 95.20, and Nano-Banana climbs to 97.05. The highest ITC score of {99.12} is achieved by {Gemini~2.5~Pro (\texttt{MAGMA-Edu})}, demonstrating its superior multimodal integration and cross-modal understanding ability.

\textbf{Overall observation.}
\texttt{MAGMA-Edu} proves to be a universally effective enhancement approach, yielding positive gains for all backbone models. Among them, {Gemini~2.5~Pro (\texttt{MAGMA-Edu})} shows the most balanced and outstanding performance in both textual and visual aspects, highlighting its robustness and generalization ability in complex multimodal educational tasks.

\begin{figure*}[htbp]
  \centering % 整体居中对齐
  % 第一个子图：宽度占单栏的48%（预留2%间距）
  \begin{subfigure}[b]{0.48\textwidth}
    \centering
    % 插入图片：width=\linewidth 让图片自适应子图宽度
    \includegraphics[width=\linewidth]{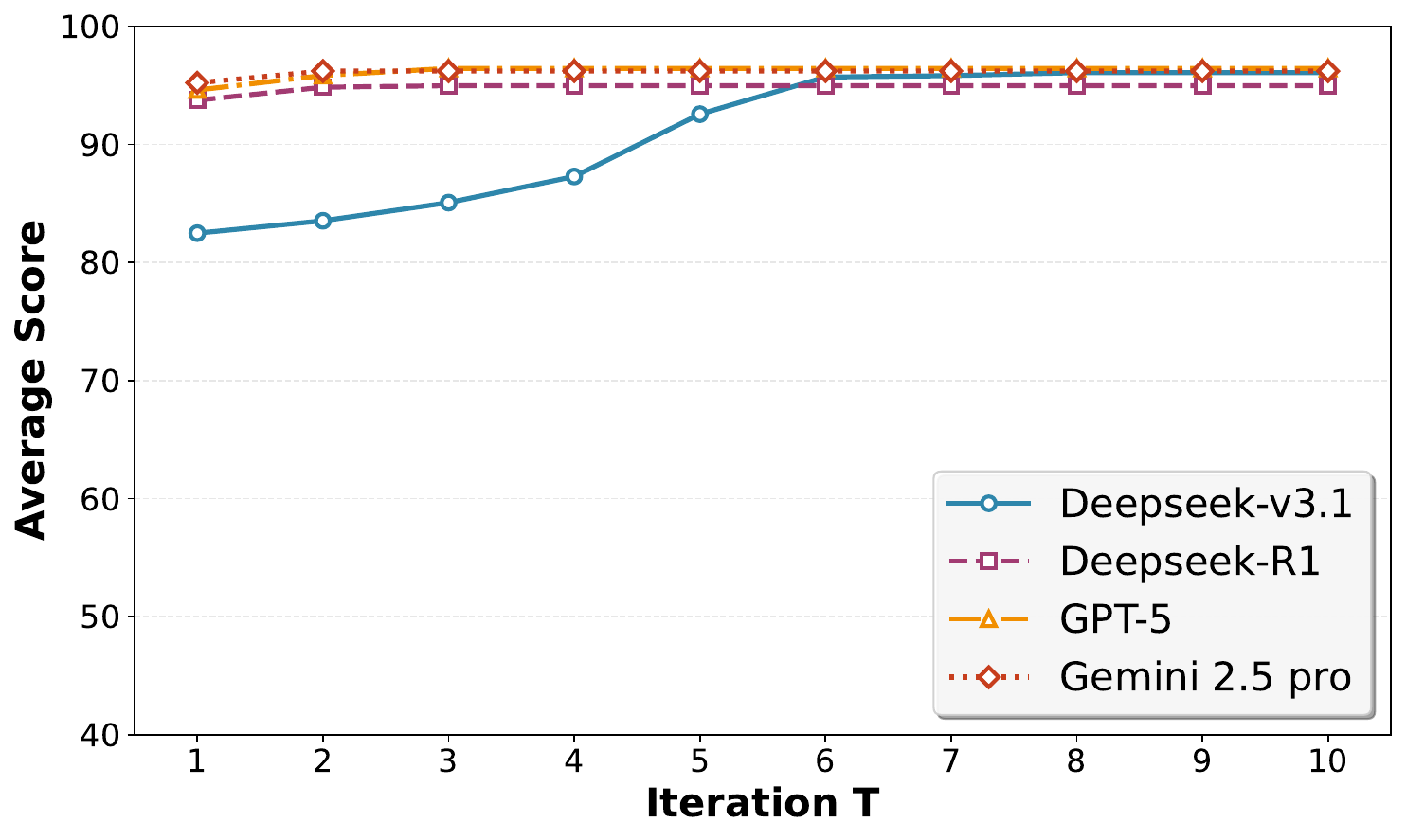} % 替换为你的图片路径/文件名
    \subcaption{Stage 1: Text Generation} % 子图小标题
    \label{Stage1} % 子图标签（用于引用）
  \end{subfigure}
  \hfill % 两个子图之间水平填充（自动分配间距）
  % 第二个子图：与第一个宽度一致（48%单栏宽度）
  \begin{subfigure}[b]{0.48\textwidth}
    \centering
    \includegraphics[width=\linewidth]{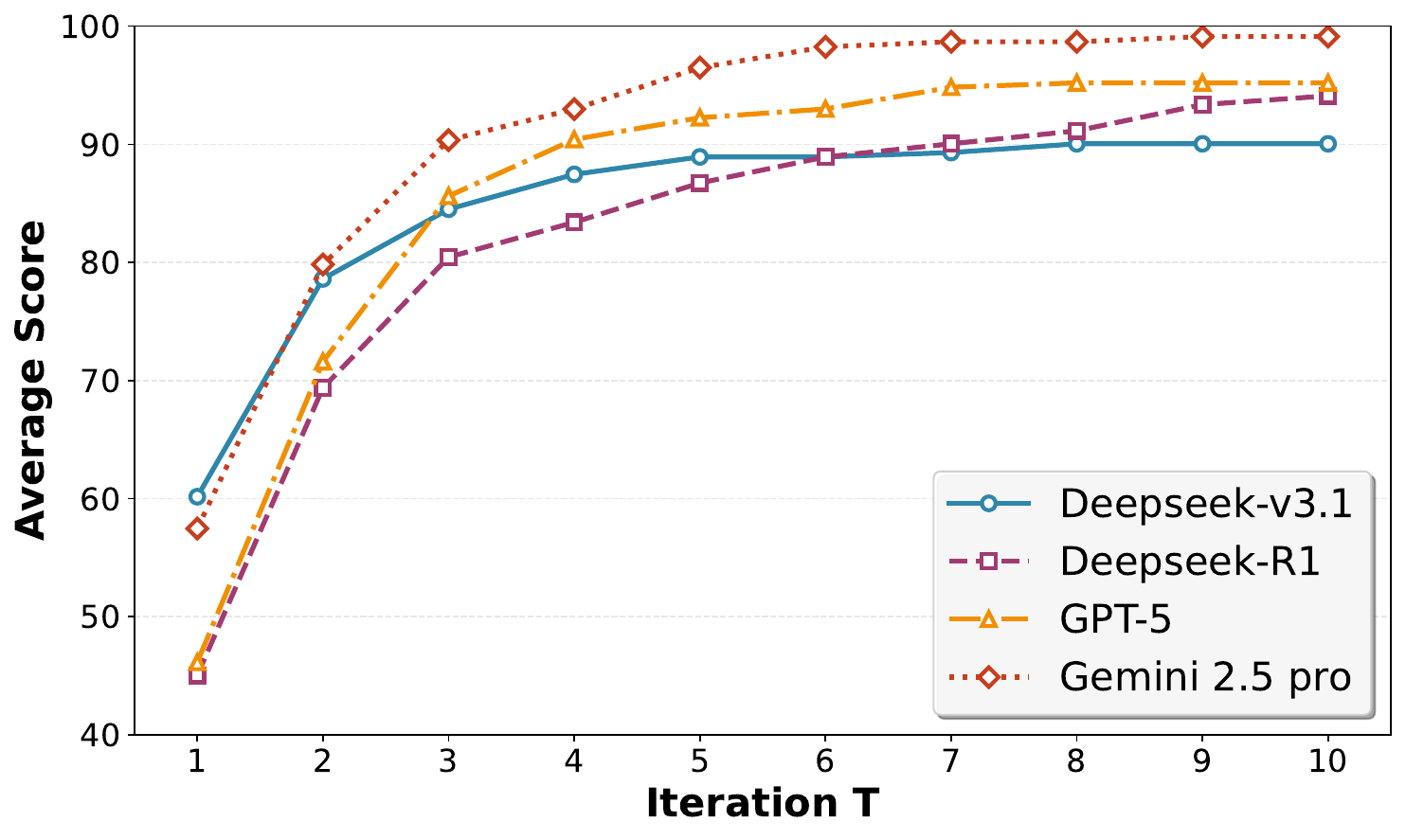} % 替换为你的图片路径/文件名
    \subcaption{Stage 2: Image Generation} % 子图小标题
    \label{Stage2} % 子图标签（用于引用）
  \end{subfigure}
  
  \caption{Effect of reflection frequency on model performance in Stage 1 and Stage 2.
The average score of Stage 1 is computed as the mean of six metrics (UO, LR, QF, AA, CA, and IDQ), while Stage 2 uses the ITC metric.} % 整图大标题
  \label{fig:reflective} % 整图标签（用于引用）
\end{figure*}

\begin{table*}[!t]
\centering
\caption{Performance comparison of Multimodal LLMs and LLMs enhanced with the MAGMA‑Edu framework across different geometry and function knowledge points.}
\label{tab:geometry_knowledge}
\setlength{\tabcolsep}{6pt}
\renewcommand{\arraystretch}{1.1}

\begin{adjustbox}{max width=\textwidth}
\begin{tabular}{llccccc}
\toprule
\textbf{Category} & \textbf{Model} 
& \textbf{Plane Geometry} 
& \textbf{Spatial Geometry} 
& \textbf{Function Images} 
& \textbf{Analytic Geometry} 
& \textbf{Mixed Knowledge} \\
\midrule

% ---------- Multimodal LLMs ----------
\multirow{2}{*}{\textbf{MLLMs}} 
 & GPT‑4o‑All & 5.34 & 12.50 & 11.69 & 24.24 & 0.00 \\
 & Nano‑Banana & 16.03 & 6.25 & 10.39 & 56.25 & 0.00 \\

\midrule

% ---------- Text‑only LLMs + MAGMA Framework ----------
\multirow{3}{*}{\textbf{MAGMA‑Edu}} 
 & GPT‑5 & 96.95 & 93.75 & 93.51 & 93.94 & 82.35 \\
 & DeepSeek‑R1 & 93.89 & 93.33 & 94.67 & 96.88 & 94.12 \\
 & Gemini 2.5 Pro & 98.13 & 100.00 & 100.00 & 100.00 & 100.00 \\

\bottomrule
\end{tabular}
\end{adjustbox}
\end{table*}

\subsection{Ablation Study Discussion}

Table~\ref{tab:ablation} shows MAGMA-Edu’s ablation results, demonstrating how training stages affect textual/visual performance across models.

\textbf{Multimodal LLMs.} 
For LLMs such as GPT‑4o and Nano‑Banana, incorporating \texttt{MAGMA‑Edu} brings large visual and textual gains. 
GPT‑4o improves from 57.01 to 92.31 in Avg‑Text (+35.30) and from 13.20 to 85.24 in ITC (+72.04), 
while Nano‑Banana shows smaller textual (+0.8) but strong visual (+81.15) gains, 
indicating effective complementation of multimodal capabilities.  

\textbf{Text‑only LLMs.} 
For GPT‑5, DeepSeek‑R1, and Gemini‑2.5‑Pro, 
textual improvements are modest, but visual consistency increases sharply. 
GPT‑5 and DeepSeek‑R1 gain about +1.23 in Avg‑Text and +95.2/+94.1 in ITC, 
while Gemini‑2.5‑Pro (Avg‑Text = 96.20) reaches 99.12 in ITC (+99.12), 
demonstrating enhanced visual reasoning for text‑only models.  

% \textbf{Stage‑wise effects.} 
% Stage~1 mainly improves textual grounding (e.g., +35.3 on GPT‑4o), 
% Stage~2 strengthens multimodal alignment (e.g., +62.45 in ITC), 
% and the final model integrates both, yielding balanced textual and visual gains.

\textbf{Stage‑wise effects.} 
Each stage contributes complementary gains. 
Stage 1 enhances textual grounding, largely boosting Avg‑Text scores (e.g., +35.3 on GPT‑4o). 
Stage 2 focuses on multimodal alignment, yielding major ITC gains (e.g., +62.45 on GPT‑4o). 
Combining both, \texttt{MAGMA‑Edu} achieves balanced improvements in language and visual understanding, highlighting the interdependence of the two modalities.

% Overall, MAGMA-Edu consistently enhances textual/visual reasoning across architectures—delivering pronounced gains for weaker baselines (compensating joint representation gaps) and measurable visual benefits for advanced models, serving as a general, scalable strategy for strengthening multimodal understanding in LLMs.

\subsection{Effect of Reflective Refinement}

Figure~\ref{fig:reflective} illustrates the influence of the iteration number $T$ on the average scores of models with different baseline levels in Stage~1 and Stage~2. 
In Stage 1 (Figure \ref{fig:reflective}a), all models show steady gains in averaged textual metrics with increased iterations. Weaker models (e.g., DeepSeek‑v3.1) benefit most from early educational feedback, while stronger ones (e.g., GPT‑5, Gemini 2.5 Pro) converge quickly with smaller improvements—reflecting Stage 1’s role in strengthening knowledge grounding and textual consistency.

In Stage 2 (Figure \ref{fig:reflective}b), which focuses on image synthesis and multimodal alignment, ITC scores rise rapidly in early iterations and stabilize around T = 7–8. MAGMA‑Edu’s self‑verification and self‑correction enable models to detect mismatches and iteratively refine outputs, sustaining improvements in image quality and alignment across both text‑only and multimodal models.

Reflective Refinement is crucial in both stages: Stage 1 enhances linguistic reasoning and conceptual grounding, while Stage 2 drives visual accuracy through iterative self‑refinement. The steady gains and convergence trends affirm MAGMA‑Edu’s robustness and scalability in advancing both textual and visual capabilities of large language models.

% Figures \ref{Stage1} and \ref{Stage2} illustrate the variation of average scores of models (deepseek-v3.2, gpt5, gemini2.5-pro, and deepseek-r1) in Stage 1 and Stage 2, respectively. \texttt{MAGMA-Edu} demonstrates strong compatibility with various models, enabling steady improvement of indicators even for state-of-the-art models. The objective of Stage 1 is to enhance the quality of question texts. Higher-performance models do not significantly raise the upper limit of the average score of multimodal texts, but they reduce the number of iterations ($t$) required to reach this limit. For instance, gemini2.5-pro basically tends to stabilize at $t=2$, while deepseek-chat does not approach the upper limit until $t=6$. Stage 2 verifies the effectiveness of our verification and reflection loop: as shown in the figures, models are not proficient in generating image code based on text initially. However, through the iterative cycle—where the verifier identifies errors in images and the reflector provides optimization suggestions—the average image quality in Stage 2 is ultimately improved from 50\% to over 90\%.

\subsection{Analysis on Knowledge Points}

As shown in Table \ref{tab:geometry_knowledge}, multimodal baselines (GPT‑4o‑All and Nano‑Banana) exhibit low and unstable accuracy across knowledge types, particularly on abstract functional and mixed problems, revealing their difficulty in linking visual understanding with symbolic reasoning.

In contrast, all LLMs enhanced by MAGMA‑Edu achieve consistently high accuracy across categories—averaging above 90 and reaching near‑perfect scores for Gemini 2.5 Pro.
The strong results of GPT‑5 and DeepSeek‑R1, originally text‑only models, further show that MAGMA‑Edu introduces reliable multimodal reasoning without weakening linguistic competence.

These findings demonstrate the broad applicability and stability of MAGMA‑Edu: it enables diverse language models to generalize across geometric perception and analytical reasoning through a unified multimodal educational framework.

%% file: sec/7_conclusion.tex
\section{Conclusion}
We proposed \texttt{MAGMA‑Edu}, a self‑reflective multi‑agent framework for generating pedagogically aligned educational problems with coherent text and diagrams. Through a generation–verification–reflection loop and a code‑based intermediate representation, \texttt{MAGMA‑Edu} ensures mathematical precision and structural clarity in diagram synthesis. Experiments on multimoal question generation show large gains over state‑of‑the‑art MLLMs, particularly in text–image consistency, confirming the effectiveness of self‑reflective collaboration and explicit structural reasoning for interpretable multimodal generation.
In future work, we will extend \texttt{MAGMA‑Edu} to broader STEM domains and real‑world instructional settings, enabling automated curriculum design and adaptive visual reasoning feedback. We also plan to explore deeper integration of symbolic reasoning with neural generation to further enhance the fidelity and explainability of educational content.

% MAGMA-Edu is a self-reflective multi-agent framework for generating pedagogically aligned educational problems with coherent text-visual components. Integrating a generation–verification–reflection loop and code-based intermediate representation, it achieves mathematical accuracy and structural clarity in diagram construction. Experiments on multimodal educational benchmarks show significant improvements over SOTA MLLMs, with large gains in text-image consistency—validating the effectiveness of self-reflective collaboration and explicit structural reasoning for reliable, interpretable multimodal generation.
% We plan to extend MAGMA-Edu to broader STEM domains and real-world instructional scenarios, supporting automatic curriculum design and adaptive visual reasoning feedback. We also aim to explore closer integration of symbolic reasoning and neural generation to enhance the fidelity and explainability of educational content from large vision-language models.

%% file: sec/X_suppl.tex
\clearpage
\appendix
\setcounter{page}{1}
\maketitlesupplementary

\section{Agent}
\subsection{Text Generator Agent}
\begin{lstlisting}
system_message:
You are a professional question generation assistant. Generate standard questions, explanations, and answers based on user needs; if images are required, additionally generate detailed image descriptions (clearly specifying image type, data, layout, etc.).
prompt:
Please generate questions, explanations, answers, and image descriptions based on user needs. The image description refers to the image required for the question. Ensure that the conditions expressed in the image description are consistent with the text conditions. Output strictly in the following format (no additional content) and do not generate images:
[Question]<Question content>
[Explanation]<Explanation content>
[Answer]<Answer content>
[Image Description]<Image description content>
The image description must complete the following tasks: (1) Describe all basic elements of the figure, including which lines, angles, points, surfaces, shapes, etc., and clearly specify the specific shapes. (2) The relative relationships of these basic elements, including positional relationships (left, right, up, down) and connection methods (intersecting, parallel, perpendicular, tangent, etc.). (3) Clearly state the numerical values of each element, such as the length of line segments, angles, etc.
User input: 
{user_input}
\end{lstlisting}

\subsection{Text Validator Agent}
\begin{lstlisting}
prompt1:
Determine whether the knowledge points and question type of the question meet the requirements of the user input. It is considered compliant if the knowledge points are involved, and the question type is compliant as long as it is correct. Output requirements: Provide verification results and reasons. Do not repeat the question or add extra explanations. Be concise and clear.

prompt2:
Determine whether the language of the question, analysis, and answer is smooth, and whether the use of symbols is correct (no grammatical errors, no format confusion). Output requirements: Provide verification results and reasons. Do not repeat the question or add extra explanations. Be concise and clear.

prompt3:
Determine whether the question and image description are clear, the conditions are complete, and whether it can be solved normally (no ambiguity, no missing key conditions). If there is a conflict between the question information and the image information, it is deemed incorrect by default. Output requirements: Provide verification results and reasons. Do not repeat the question or add extra explanations. Be concise and clear.

prompt4:
Verification task: Determine whether the analysis is correct, the steps are complete, and whether it can accurately solve the question. Output requirements: Provide verification results and reasons. Do not repeat the question or add extra explanations. Be concise and clear.

prompt5:
Please judge whether the answer is correct based on the question and image description. Output requirements: Provide verification results and reasons. Do not repeat the question or add extra explanations. Be concise and clear.

prompt6:
The image description must complete the following tasks: 
(1) Describe all basic elements of the figure, including which lines, angles, points, surfaces, shapes, etc., and clearly specify the specific shapes.
(2) The relative relationships of these basic elements, including positional relationships (left, right, up, down) and connection methods (intersecting, parallel, perpendicular, tangent, etc.).
(3) Clearly state the numerical values of each element, such as the length of line segments, angles, etc. Output requirements: Provide verification results (passed or not passed) and reasons. Do not repeat the question or add extra explanations. Be concise and clear.
\end{lstlisting}

\subsection{Text Reflector Agent}
\begin{lstlisting}
prompt: 
Based on all fields of the original question and all verification results, determine whether the question and image description need to be revised. Summarize the parts that need revision and provide revision suggestions. Note that if there is a conflict between the question information and the image description information, prioritize revising the image description rather than the question. Output strictly in the following format:
[Question]<Question content>
[Explanation]<Explanation content>
[Answer]<Answer content>
[Image Description]<Image description content>
\end{lstlisting}

\subsection{Code Generator Agent }
\begin{lstlisting}
prompt:
Generate directly executable Python image code based on the image description, strictly following the following requirements:
1. Must use the matplotlib library (preferred) or seaborn library; other plotting libraries are prohibited.
2. The code must be complete and executable, including all necessary imports, data definitions, plotting logic, and saving steps.
3. The fixed image saving path is: '{image_save_path}' (the path has been automatically created, please use it directly).
4. Generate strictly according to the image description:
   - Must include all basic images such as all figures, points, lines, surfaces, angles mentioned in the image description.
   - Must draw according to the positional relationships of the figures required by the image description.
   - Must correctly label numerical values such as line segment lengths and angles, corresponding to the objects to be labeled in the question.
   - Do not generate any special symbols, such as parallel symbols, perpendicular symbols, etc.
5. Code format specifications:
   - Indent with 4 spaces, add clear comments, do not use any Chinese symbols. Except as required by the image description, comment on the coordinate system.
   - Avoid redundant code and ensure no syntax errors.
   - Do not wrap with ```python```, output the code content directly.
6. Only output executable Python code, no additional explanations or markdown formats.
\end{lstlisting}

\subsection{Code Executor Agent }
\begin{lstlisting}
prompt: Be responsible for executing the generated Python image code, capture the execution result (success/error), and return it to the verification Agent.
\end{lstlisting}

\subsection{Image Validator Agent}
\begin{lstlisting}
prompt1:
Based on the image description, generated code, and code execution result, comprehensively judge whether the code meets the requirements. Evaluation criteria:
1. Pass if all elements in the image description exist and their positional relationships are correct.
2. Pass if the required standard numerical labels exist; incorrect labeling of angle symbols, right angle symbols, perpendicular symbols, etc., is not considered an error.
3. Pass if there is image overlap.
4. Pass if there are no special marks (e.g., parallel symbols, perpendicular symbols, angle symbols, etc.).

Output requirements:
- First clearly state the verification result (Pass/Fail);
- Then explain the reason (focus on pointing out issues: execution error/missing elements/inconsistency, etc.);
- If failing, provide specific modification directions; directly return "Pass" if partially passing;
- Control the total number of words within 150, no redundancy.
\end{lstlisting}

\begin{lstlisting}
prompt2: 
As a multimodal verification tool, compare the consistency between the generated image and the original description to determine if there are errors.
Input: Image (base64 format), original image description
Verification criteria:
1. Correct if all elements in the image description exist and their positional relationships are correct.
2. Correct if the required standard numerical labels exist; incorrect positions of angle symbols, right angle symbols, perpendicular symbols, etc., are not considered errors.
3. Missing points/lines are not considered errors if they may be due to image overlap.
4. Absence of special marks (e.g., parallel symbols, perpendicular symbols, angle symbols, etc.) is not considered an error.

Output requirements:
- First clearly state the verification result (Pass/Fail);
- Briefly explain the basis (1-2 sentences);
- If failing, clearly point out differences between the image and description and modification suggestions; directly return "Pass" if partially passing;
- Control the total number of words within 120, no additional explanations.
\end{lstlisting}

\subsection{Image Reflector Agent}
\begin{lstlisting}
 I am generating image code. The previously generated code and verification results are as follows:
1. Previously generated code: {cycle_code}
2. Code execution result: {last_exec_result[:100]}...
3. Large model verification result: {last_code_val_result}
4. Multimodal verification result: {last_multimodal_val_result}

Please optimize the code based on the above modification suggestions, with the following requirements:
- Strictly iterate based on the previous code: prioritize fixing code errors, then optimize inconsistencies between the code and the image. If there is overlap (e.g., two coincident points), adjust the code (e.g., express one point in parentheses); do not modify error-free parts.
- Ensure compliance with the original image description: {image_description};
- Code requirements:
  - Must include all basic elements (figures, points, lines, surfaces, angles, etc.) mentioned in the image description.
  - Must draw according to the positional relationships of figures required by the image description.
  - Must correctly label numerical values (e.g., line segment lengths, angles) corresponding to the objects to be labeled in the question.
  - Do not generate any special symbols (e.g., parallel symbols, perpendicular symbols, angle symbols, etc.).
- The save path remains: {image_save_path};
- The code must be complete, executable, and free of redundancy.
\end{lstlisting}

\section{Criteria}

Evaluation Criteria Proposed by Zhou \cite{zhou2025answers}:
\begin{itemize}[leftmargin=1.5cm, itemsep=10pt] % 黑点列表，调整间距与缩进
    \item \textbf{Knowledge Point Alignment (KP)}
    Determine whether the generated questions accurately and comprehensively cover the knowledge points specified by the user, avoiding deviation from the theme or interdisciplinary content.
    
    \item \textbf{Question Type Alignment (QT)}
    The question type must be consistent with the user's requirements (e.g., multiple-choice questions, fill-in-the-blank questions, problem-solving questions) and comply with the standard format of the type (e.g., multiple-choice questions include four options, fill-in-the-blank questions have clear answer indicators).
    
    \item \textbf{Question Quality (QQ)}
    The questions are expressed clearly and concisely with standard terminology, featuring clear and solvable assessment objectives. They are free of ambiguity, logical fallacies, or typos, facilitating students' understanding of the question intent.
    
    \item \textbf{Solution Quality (SQ)}
    The solution process is correct, rigorous, and complete. The involved knowledge points are compatible with the curriculum requirements of the target academic stage, and the correct answer can be derived through the explanation without omitting key steps.
    
    \item \textbf{Competency-Guided (CG)}
    Questions should integrate real scenarios (e.g., daily life applications, subject-specific practices) to guide students in applying knowledge to develop higher-order cognitive abilities, avoiding mere assessment of purely abstract knowledge points.
\end{itemize}

Evaluation Criteria Proposed by ours:
\begin{itemize}[leftmargin=1.5cm, itemsep=10pt]
    \item \textbf{User Orientation (UO)} — Evaluates whether the generated question satisfies the instructional requirements specified in the system input $\mathcal{I}$.
    \item \textbf{Language Readability (LR)} — Ensures grammatical fluency and the absence of corrupted characters or non‑standard symbols in the generated content.
    \item \textbf{Question Feasibility (QF)} — Assesses the rationality and pedagogical appropriateness of the question stem and its associated image information.
    \item \textbf{Accurate Analysis (AA)} — Examines the logical soundness and coherence of the reasoning presented in the generated explanation or solution.
    \item \textbf{Correct Answer (CA)} — Verifies the numerical or symbolic correctness of the final answer derived for the question.
    \item \textbf{Image Description Quality (IDQ)} — Evaluates whether the image description accurately captures the intended textual requirements and visual context.
    \item \textbf{Image–Text Consistency (ITC)} — A question is considered valid if it successfully passes all three verification stages: code quality $\{Q_{\text{syntax}}\}$, code‑text alignment $\{Q_{\text{align}}\}$, 
    and multimodal reasoning $\{Q_{\text{visual}}\}$.
\end{itemize}

\textbf{Adjustments to Core Dimensions}:
\begin{itemize}[leftmargin=1.5cm, itemsep=10pt]
    \item For the knowledge point and question type dimensions, we maintain consistency with the standards proposed by Zhou, but merge the two into a single evaluation item.
    \item Regarding the question quality dimension, additional consideration of image adaptability is required in the multimodal scenario, leading to differences between our standards and the original version.
    \item The solution explanation quality dimension is retained. Meanwhile, combined with testing practice, a new "answer quality" assessment is added — in practical applications, it was found that not only explanations may contain errors, but answers themselves can also be flawed.
\end{itemize}

\textbf{New Key Evaluation Items}:
\begin{itemize}[leftmargin=1.5cm, itemsep=10pt]
    \item Image description evaluation: One of our core processes involves generating code based on image text descriptions and then generating images. Therefore, the completeness, conciseness, and model understandability of image descriptions are included as important evaluation criteria.
    \item Language expression evaluation: A key focus in the text domain is language standardization, which requires detecting irregular expressions, format errors, and other related issues.
    \item Image-text consistency evaluation: This is crucial for images to convey key information. We have designed a triple verification mechanism that detects from the code perspective, code-text perspective, and image-text perspective respectively to ensure high consistency between images and text descriptions.
\end{itemize}

\section{sample data}
multimodal prompts:
\begin{lstlisting}
grade: "Seventh Grade of Junior High School"
knowledge_point: "Properties of Complementary Angles and Supplementary Angles"
instruction: "A seventh-grade junior high school math teacher is designing questions for the upcoming unit test. They need an application question to assess students' understanding of the properties of complementary angles and supplementary angles. The question must include a figure, require students to solve it by combining the figure, and be at a basic level."
\end{lstlisting}

Gemini 2.5 Pro generated examples:
\begin{lstlisting}[language=TeX, caption={Complementary and Supplementary Angles Application Problem}, inputencoding=utf8]
user_input:
A seventh-grade junior high school math teacher is designing questions for an upcoming unit test. They need an application problem that assesses students' understanding of the properties of complementary and supplementary angles. The question must include a graph, require students to solve it by combining the graph, and be at a basic difficulty level.

question:
As shown in the figure, points A, O, and B are collinear, with OC perpendicular to AB at point O. OD is a ray inside angle BOC. If angle BOD = 30 degrees, find the measures of angle COD and angle AOD.

analysis:
1. First, according to the problem condition "OC vertical AB", we know that angle BOC is a right angle, so angle BOC = 90 degrees.
2. Since OD is a ray inside angle BOC, angle COD and angle BOD are complementary angles. Their sum equals angle BOC. Therefore, the measure of angle COD can be found by subtracting the known measure of angle BOD from angle BOC:
   angle COD = angle BOC - angle BOD = 90 degrees - 30 degrees = 60 degrees.
3. Next, according to the problem condition "points A, O, and B are collinear", we know that angle AOB is a straight angle, so angle AOB = 180 degrees.
4. Since angle AOD and angle BOD are supplementary angles, their sum equals angle AOB. Therefore, the measure of angle AOD can be found by subtracting the known measure of angle BOD from angle AOB:
   angle AOD = angle AOB - angle BOD = 180 degrees - 30 degrees = 150 degrees.
   (Alternative method: angle AOD = angle AOC + angle COD. Since OC vertical AB, angle AOC = 90 degrees. From the previous step, we have found angle COD = 60 degrees, so angle AOD = 90 degrees + 60 degrees = 150 degrees.)

answer:
angle COD = 60 degrees, angle AOD = 150 degrees.

image_description:
A horizontal straight line AOB, with point O between points A and B. A ray OC extends vertically upward from point O, and OC vertical AB, thus forming two right angles of 90 degrees: angle AOC and angle BOC. Another ray OD is inside the right angle BOC, starting from point O. Its position makes angle BOD = 30 degrees. Since angle BOC = 90 degrees, ray OD is closer to ray OB. The angles to be found in the figure are angle COD (between rays OC and OD) and angle AOD (between rays OA and OD).
plot_code:
import matplotlib.pyplot as plt
import numpy as np
fig, ax = plt.subplots(figsize=(8, 6))
ax.set_xlim(-5, 5)
ax.set_ylim(-1, 5)
ax.set_aspect('equal')
ax.axis('off')
O = (0, 0)
A = (-4, 0)
B = (4, 0)
C = (0, 4)
ax.plot([A[0], B[0]], [A[1], B[1]], 'k-', linewidth=1.5)
ax.plot([O[0], C[0]], [O[1], C[1]], 'k-', linewidth=1.5)
angle_BOD_deg = 30
angle_BOD_rad = np.deg2rad(angle_BOD_deg)
ray_length = 4
D_x = ray_length * np.cos(angle_BOD_rad)
D_y = ray_length * np.sin(angle_BOD_rad)
D = (D_x, D_y)
ax.plot([O[0], D[0]], [O[1], D[1]], 'k-', linewidth=1.5)
ax.plot(A[0], A[1], 'ko', markersize=3)
ax.plot(O[0], O[1], 'ko', markersize=3)
ax.plot(B[0], B[1], 'ko', markersize=3)
ax.plot(C[0], C[1], 'ko', markersize=3)
ax.plot(D[0], D[1], 'ko', markersize=3)
ax.text(A[0] - 0.2, A[1] - 0.2, 'A', fontsize=12)
ax.text(O[0] - 0.2, O[1] - 0.2, 'O', fontsize=12)
ax.text(B[0] + 0.1, B[1] - 0.2, 'B', fontsize=12)
ax.text(C[0] + 0.1, C[1], 'C', fontsize=12)
ax.text(D[0] + 0.1, D[1] + 0.1, 'D', fontsize=12)
def draw_angle_arc(center, radius, start_angle, end_angle, label, label_pos_radius, label_offset_angle):
    theta = np.linspace(np.deg2rad(start_angle), np.deg2rad(end_angle), 100)
    x_arc = center[0] + radius * np.cos(theta)
    y_arc = center[1] + radius * np.sin(theta)
    ax.plot(x_arc, y_arc, 'k--', linewidth=0.8)
    mid_angle = np.deg2rad((start_angle + end_angle) / 2 + label_offset_angle)
    label_x = center[0] + label_pos_radius * np.cos(mid_angle)
    label_y = center[1] + label_pos_radius * np.sin(mid_angle)
    ax.text(label_x, label_y, label, fontsize=10, ha='center', va='center')
rect_size = 0.3
ax.plot([O[0], O[0], O[0] - rect_size, O[0] - rect_size], 
        [O[1] + rect_size, O[1], O[1], O[1] + rect_size], 'k-', linewidth=0.8)
draw_angle_arc(O, 1.5, 90, 180, '90 degrees', 1.8, 0)
ax.plot([O[0], O[0] + rect_size, O[0] + rect_size, O[0]], 
        [O[1] + rect_size, O[1] + rect_size, O[1], O[1]], 'k-', linewidth=0.8)
draw_angle_arc(O, 1.5, 0, 90, '90 degrees', 1.8, 0)
draw_angle_arc(O, 1.0, 0, 30, '30 degrees', 1.2, 0)
angle_COD_deg = 90 - angle_BOD_deg
angle_AOD_deg = 180 - angle_BOD_deg
draw_angle_arc(O, 2.5, 30, 90, 'angle COD', 2.8, 0)
ax.text(0.5, 3.0, '60 degrees', fontsize=10, ha='center', va='center', color='blue')  # Label value for COD
draw_angle_arc(O, 3.0, 30, 180, 'angle AOD', 3.3, 0)
ax.text(-2.0, 1.5, '150 degrees', fontsize=10, ha='center', va='center', color='red')  # Label value for AOD
plt.savefig('img/image_0.png')
plt.close()
\end{lstlisting}

\section{Real Examples}
\begin{figure}
    \centering
    % 总标题（整个图的统一说明，会自动分配唯一编号，如 Fig. 1）
    \caption{Comparison between images generated based on nano-banana and those generated by code}
    \label{fig:two-methods-with-sublabels}
    
    % 图a：Code-based 子图
    \begin{subfigure}[t]{0.45\linewidth}
        \centering
        \includegraphics[width=\linewidth]{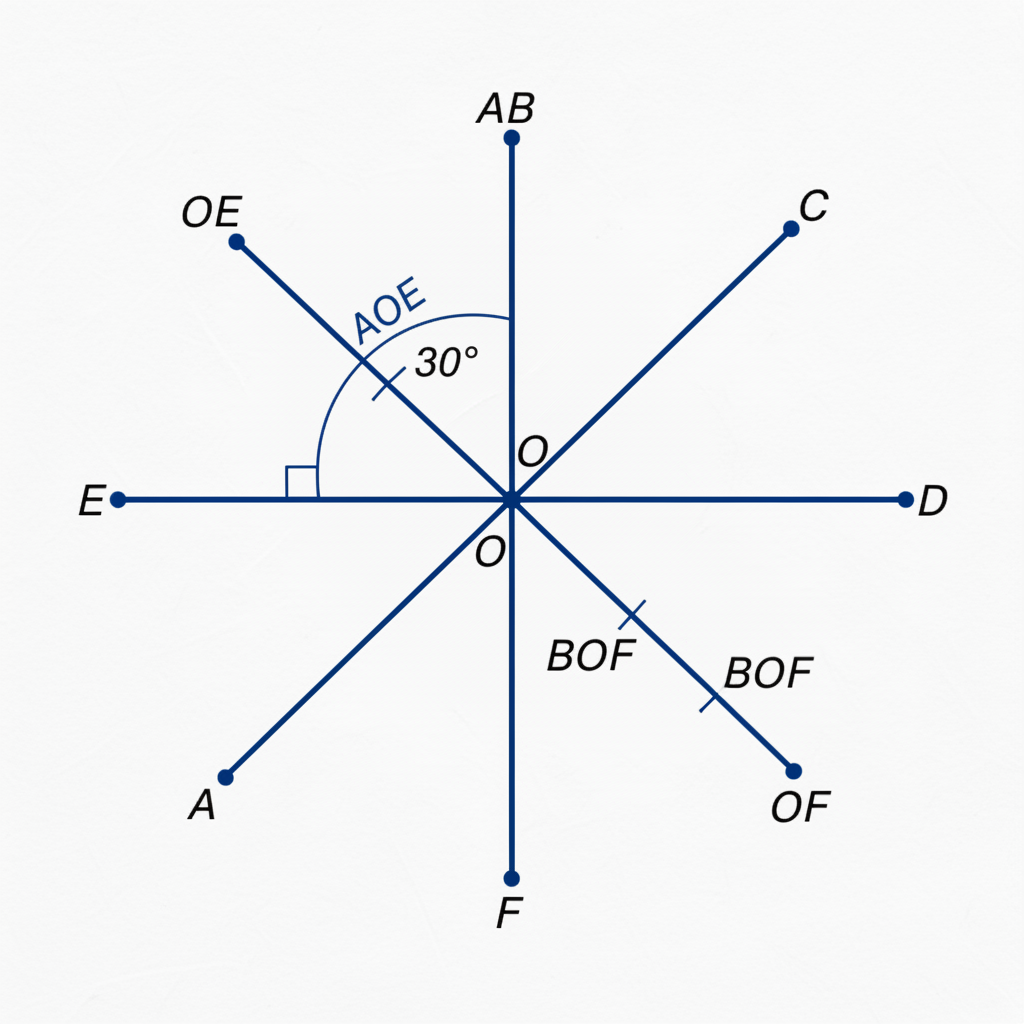}
        \subcaption{Base on Code}  % 子标签说明（对应 (a)）
        \label{fig:sub-code}  % 可选：子图独立标签（如需单独引用）
    \end{subfigure}
    \hfill  % 自动填充两图间距，保持居中对齐
    % 图b：LLM-based 子图
    \begin{subfigure}[t]{0.45\linewidth}
        \centering
        \includegraphics[width=\linewidth]{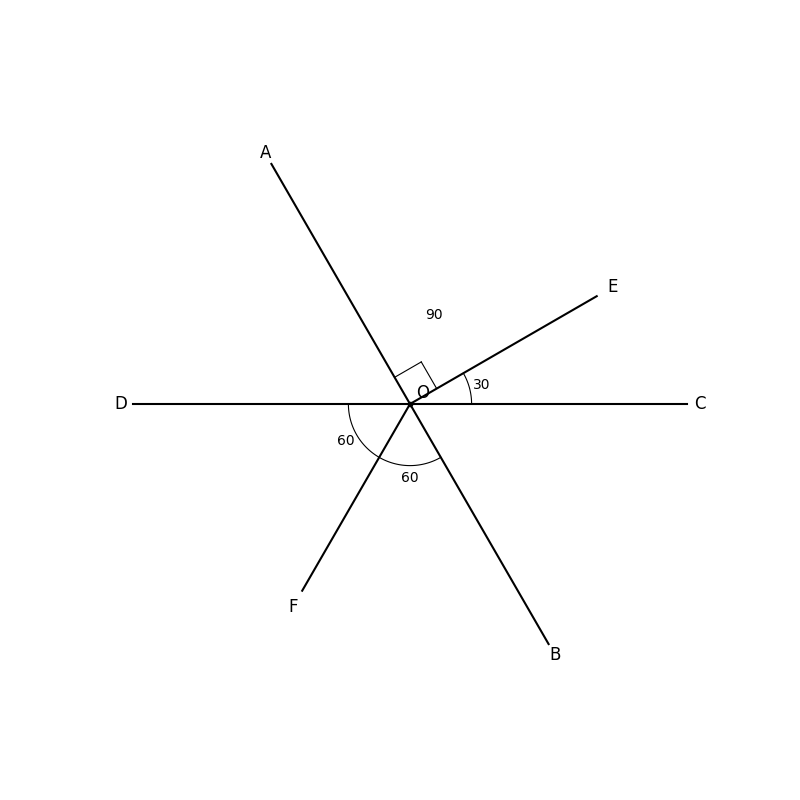}
        \subcaption{Base on LLM}  % 子标签说明（对应 (b)）
        \label{fig:sub-llm}  % 可选：子图独立标签
    \end{subfigure}
\end{figure}

Comparison between images generated based on nano-banana and those generated by code, as shown in the Figure \ref{fig:two-methods-with-sublabels}:

\textbf{Image Description}:
Within a plane, there are two straight lines $AB$ and $CD$ that intersect at point $O$. Ray $OE$ is drawn from point $O$, lies inside angle $AOC$, and angle $AOE = ninety degrees$. Ray $OF$ is drawn from point $O$ and bisects angle $BOD$. It is known that angle $COE = thirty degrees$. (The image should clearly show that angle $AOE$ is a right angle, and $OE$ lies inside angle $AOC$ such that angle $AOC$ is clearly greater than ninety degrees.)
% \textbf{Model generated images}:
% \begin{figure}
%     \centering
%     \includegraphics[width=0.5\linewidth]{figures/LM.png}
%     \caption{Base on LLM}
%     \label{fig:placeholder}
% \end{figure}
% \textbf{Ours Mothed}:
% \begin{figure}
%     \centering
%     \includegraphics[width=0.5\linewidth]{figures/code.png}
%     \caption{Base on Code}
%     \label{fig:placeholder}
% \end{figure}

Under the same description conditions, intuitively, the images generated by the code-based method can basically meet the requirements. The elements mentioned in the image description—such as line segments AB, CD, OE, OF, and angles—are all displayed. However, the LLM-based method has issues: for example, the line segments AB and CD are not labeled at all; instead, line segments like AC and DE appear incorrectly. Problems also exist in angle labeling—for instance, it is difficult to identify the specific angles corresponding to angle AOE and angle AOC.